\documentclass{article}
    \PassOptionsToPackage{numbers, compress}{natbib}

    \usepackage[preprint]{neurips_2024}

\usepackage{natbib}

\usepackage[table,dvipsnames]{xcolor}
\usepackage[utf8]{inputenc} 
\usepackage[T1]{fontenc}    
\usepackage{hyperref}       
\usepackage{url}            
\usepackage{booktabs}       
\usepackage{amsfonts}       
\usepackage{nicefrac}       
\usepackage{microtype}      
\usepackage{xcolor}         
\usepackage{amsmath}
\usepackage{wrapfig}
\usepackage{enumitem}
\usepackage{bm}
\usepackage{multirow}
\usepackage{amssymb}
\usepackage{algorithm}
\usepackage{algpseudocode}

\usepackage{float}
\usepackage[inkscapeformat=png]{svg}

\usepackage{pifont}
\newcommand{\cmark}{\ding{51}}%
\newcommand{\xmark}{\ding{55}}%
\definecolor{applegreen}{rgb}{0.55, 0.71, 0.0}
\title{Compositional Video Generation as Flow Equalization}

\author{
    Xingyi Yang \quad 
    Xinchao Wang\thanks{Corresponding Author} \\
    National University of Singapore \\
    \texttt{xyang@u.nus.edu \quad xinchao@nus.edu.sg} \\
}

\begin{document}

\maketitle

\begin{figure}[h]
\vspace{-6mm}
    \centering
    \includegraphics[width=\linewidth]{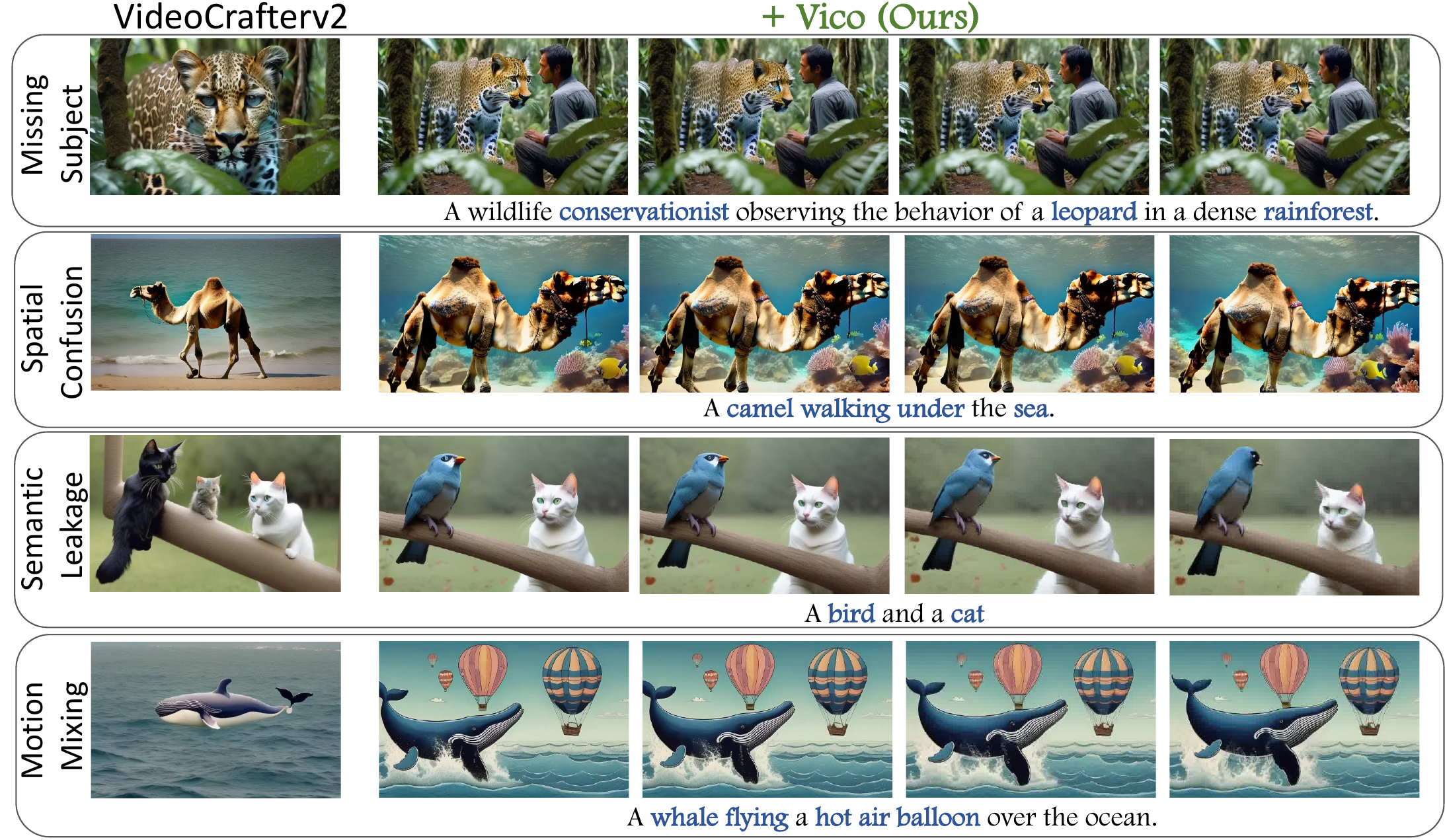}
    \vspace{-6mm}
    \caption{Examples for compositional video generation of  \textbf{Vico} on top of VideCrafterv2~\cite{chen2024videocrafter2}. We identify four types of typical failure in compositional T2V~(Row 1) \textit{Missing Subject} (Row 2) \textit{Spatial
Confusion} (Row 3) \textit{Semantic Leakage} and (Row 4) \textit{Motion Mixing}. \textbf{Vico} provides a unified solution to these issues by equalizing the contributions of text tokens.}
    \label{fig:examples}
\end{figure}
\begin{abstract}
    Large-scale Text-to-Video (T2V) diffusion models 
    have recently demonstrated unprecedented
    capability to transform natural language descriptions into stunning and photorealistic videos. 
    Despite the promising results, a significant challenge remains: 
    these models struggle to fully grasp complex compositional 
    interactions between multiple concepts and actions. {This issue arises when some words dominantly influence the final video, overshadowing other concepts.}
    To tackle this problem,     
    we introduce \textbf{Vico}, a generic framework for compositional video generation 
    that explicitly ensures all concepts are represented properly.
At its core, Vico analyzes how input tokens influence the generated video, 
and adjusts the model to prevent any single concept from dominating. 
Specifically, Vico extracts attention weights from all layers to build a spatial-temporal attention graph, and then estimates the influence
as the \emph{max-flow} from the source text token to the video target token.
Although the direct computation of attention flow in diffusion models is typically infeasible,
we devise an efficient approximation based on
subgraph flows and employ 
a fast and vectorized implementation,
which in turn makes the flow computation
manageable and differentiable.
By updating the noisy latent to balance these flows, Vico captures complex interactions and consequently produces videos that closely adhere to textual descriptions. We apply our method to multiple diffusion-based video models for compositional T2V and video editing. Empirical results demonstrate that our framework significantly enhances the compositional richness and accuracy of the generated videos. Visit our website at~\href{https://adamdad.github.io/vico/}{\url{https://adamdad.github.io/vico/}}.

    
\end{abstract}

\section{Introduction}

Humans recognize the world compositionally. That is to say, we perceive and understand the world by identifying parts of objects and assembling them into a whole. This ability to recognize and recombine elements—making ``infinite use of finite mean''—is crucial for understanding and modeling our environment. Similarly, in the realm of generative AI, particularly in video generation, it is crucial to replicate this compositional approach.

Despite advancements in generative models, current diffusion models fail to capture the true compositional nature of inputs. Typically, some words disproportionately influence the generative process, leading to visual content that does not reflect the intended composition of elements. While the compositional text-to-image sythesis~\cite{liu2022compositional,10.1145/3592116,kumari2022customdiffusion,feng2023trainingfree,huang2023t2i} has been more studied, the challenge of compositional video generation has received less attention. This oversight is largely due to the high-dimensional nature of video and the complex interplay between concepts and motion. 

As an illustration, we highlight some failure cases in Figure~\ref{fig:examples}~(Left), where \emph{certain words dominate} while others are underrepresented. Common issues include \emph{missing subject} and \emph{spatial confusion}, where some concepts do not appear in the video. Even with all concepts present, \emph{semantic leakage} can occur, causing attributes amplified incorrectly, for example, \emph{a bird looks like a cat}. A challenge specific to T2V is \emph{Motion Mixing}, where the action intended for one object mistakenly interacts with another, such as generating a \texttt{flying wale} instead of \texttt{flying balloon}.


To address these challenges, we present \textbf{Vico}, a novel framework for compositional video generation 
that ensures all concepts are represented equally. 
Vico operates on the principle that, 
each textual token should have an equal opportunity 
to influence the final video output. At our core, 
Vico first assesses and then rebalances the influence of these tokens. 
This is achieved through test-time optimization, where we assess and adjust the impact of each token at every reverse time step of our video diffusion model. As shown in~\ref{fig:examples}, Vico resolves the above questions and provides better results.

One significant challenge 
is accurately attributing text influence. 
While cross-attention~\cite{tang-etal-2023-daam,mokady2022null,feng2023trainingfree,rassin2024linguistic} provides faithful attribution 
in text-to-image diffusion models, it is not well-suited for video models. It is because such cross-attention 
is only applied on spatial modules along,
treating each frame independently, 
without directly influencing temporal dynamics. 

To surmount this, we develop
a new attribution method for T2V model, termed \emph{Spatial-Temporal Attention Flow}~(\textbf{ST-flow}). ST-flow considers all attention layers of the diffusion model, and views it as a spatiotemporal flow graph. Using the maximum flow algorithm, it computes the flow values, from input tokens (sources) to video tokens (target). These values serve as our estimated contributions.

Unfortunately, this naive attention max-flow computation is, in fact, both computationally expensive and non-differentiable. 
We thus derive an efficient and differentiable approximation for the ST-Flow. Rather than computing flow values on the full graph, 
we instead compute the flow on all subgraphs. 
The ST-Flow is then
estimated as the maximum subgraph flow. 
Additionally, we have develop a special matrix operation to compute this subgraph flow in a fully vectorized manner, making it approximately 100$\times$ faster than the exact ST-flow.

Once we obtain these attribution scores, we proceed to optimize the model to balance such contributions. We do this as a min-max optimization, where we update the latent code, in the direct that, the least represented token should increase its influence.

We implement Vico on multiple video applications, including text-to-video generation and video editing. These applications highlight the framework's flexibility and effectiveness in managing complex prompt compositions, demonstrating significant improvements 
over traditional methods in both the accuracy of generated video. Our contributions can be summarized below:

\begin{itemize}
    \item We introduce \textbf{Vico}, a framework for compositional video generation. 
    It optimizes the model to ensure each input token fairly influences the final video output.
    \item We develop ST-flow, a new attribution method that uses attention max-flow to evaluate the influence of each input token in video diffusion models.
    \item We derive a differentiable method to approximate ST-flow by calculating flows within subgraphs. It greatly speed up computations with a fully vectorized implementation.
    \item Extensive evaluation of Vico in diverse settings has proven its robust capability, with substantial improvements in video quality and semantic accuracy.
\end{itemize}
\section{Preliminaries}

\noindent\textbf{Denoising Diffusion Probabilistic Models.} Diffusion model reverses a progressive noise process based on latent variables. Given data $\mathbf{x}_0 \sim q(\mathbf{x}_0)$ sampled from the real distribution, we consider perturbing data with Gaussian noise of zero mean and $\beta_t$ variance for $T$ steps/
At the end of day, $\mathbf{x}_T \to \mathcal{N}(0, \mathbf{I})$ converge to isometric Gaussian noise. The choice of Gaussian  provides a close-form solution to generate arbitrary time-step $\mathbf{x}_t$ through
\begin{align}
    \mathbf{x}_t = \sqrt{\bar{\alpha}_t} \mathbf{x}_0 + \sqrt{1-\bar{\alpha}_t} \bm{\epsilon}, \quad \text{where} \quad \bm\epsilon \sim \mathcal{N}(0, \mathbf{I})
\end{align}
where $\alpha_t = 1-\beta_t$ and $\bar{\alpha}_t = \prod_{s=1}^t \alpha_s$. A variational Markov chain in the reverse process is parameterized as a time-conditioned denoising neural network $\bm \epsilon_{\bm \theta}(\mathbf{x}, t)$ with $p_{\bm \theta}(\mathbf{x}_{t-1}|\mathbf{x}_t)=\mathcal{N}(\mathbf{x}_{t-1}; \frac{1}{\sqrt{1-\beta_t}}(\mathbf{x}_t+\beta_t \bm \epsilon_{\bm \theta}(\mathbf{x}, t)), \beta_t \mathbf{I})$. The denoiser is trained to minimize a re-weighted evidence lower bound~(ELBO) that fits the noise
\begin{align}
    \mathcal{L}_{\text{DDPM}} &= \mathbb{E}_{t,\mathbf{x}_0, \bm\epsilon} \Big[||\bm\epsilon +\sqrt{1-\Bar{\alpha}_t} \bm \epsilon_{\bm \theta}(\mathbf{x}, t)||_2^2\Big]\label{eq:loss_ddpm}
\end{align}
Training with denoising loss, $\bm \epsilon_{\bm \theta}$ equivalently learns to recover the derivative that maximize the data log-likelihood~\cite{song2019generative,hyvarinen2005estimation,vincent2011connection}.
With a trained $\bm \epsilon_{\bm \theta^*}(\mathbf{x}, t)\approx\nabla_{\mathbf{x}_t} \log p(\mathbf{x}_t)$, we generate the data by reversing the Markov chain
\begin{align}
    \mathbf{x}_{t-1}\leftarrow \frac{1}{\sqrt{1-\beta_t}}(\mathbf{x}_t+\beta_t \bm \epsilon_{\bm \theta^*}(\mathbf{x}, t)) + \sqrt{\beta_t} \bm \epsilon_t;
    \label{eq:update}
\end{align}
The reverse process could be understood as going along $\nabla_{\mathbf{x}_t} \log p(\mathbf{x}_t)$ to maximize the likelihood.

\noindent\textbf{Text-to-Video (T2V) Diffusion Models.} Given a text prompt $y$, T2V diffusion models progressively generate a video from Gaussian noise. This generation typically occurs within the latent space of an autoencoder~\cite{rombach2022high} to reduce the complexity. The architecture design of T2V models often follows either a 3D-UNet~\cite{DBLP:conf/nips/HoSGC0F22,blattmann2023align,DBLP:journals/corr/abs-2210-02303,harvey2022flexible,wu2023tune} or diffusion transformer~\cite{DBLP:journals/corr/abs-2312-06662,DBLP:conf/iccv/PeeblesX23,DBLP:journals/corr/abs-2401-03048}. For computational efficiency, these architectures commonly utilize separate self-attention~\cite{vaswani2017attention}  for spatial and temporal tokens. Moreover, cross-attentions is applied on each frame separately, thereby injecting conditions into the model. More related work is in Appendix~\ref{sec:related}.

\noindent\textbf{Maximum-Flow Problem.}~\cite{harris1955fundamentals,ford1956maximal,edmonds1972theoretical} Consider a directed graph $G(V,E)$ with a source node $s$ and a target node $t$. A flow is function on edge $f: E\to \mathbb{R}$ that satisfies both \emph{conservation constraint} and \emph{capacity constraint} at every vertex $v \in V \backslash \{s,t\}$. This means the total inflow into any node $v$ must equals its total outflow, and the flow on any edge cannot exceed its capacity. The flow value $|f|=\sum_{e_{s,v} \in E} f(s,u)$ is defined as the total flow out of the source $s$, which is equal to the total inflow into the target $t$, $|f|=\sum_{e_{u,t} \in E} f(u,t)$. The maximum flow problem is to find a flow $f^*$ that maximizes this value.

\section{Vico: Compositional Video Generation as Flow Equalization}

In this paper, we solve the problem of compositional video generation by equalizing influence among tokens. We calculate this influence using max-flow within the attention graph of the T2V model and ensure efficient computation. We define our problem and optimization scheme in Sec~\ref{sec:prob&opt}. The definition of ST-Flow and its efficient computation are discussed in Sections~\ref{sec:ST-flow} and~\ref{sec:approaximation}.
\begin{figure}
    \centering
    \includegraphics[width=\linewidth]{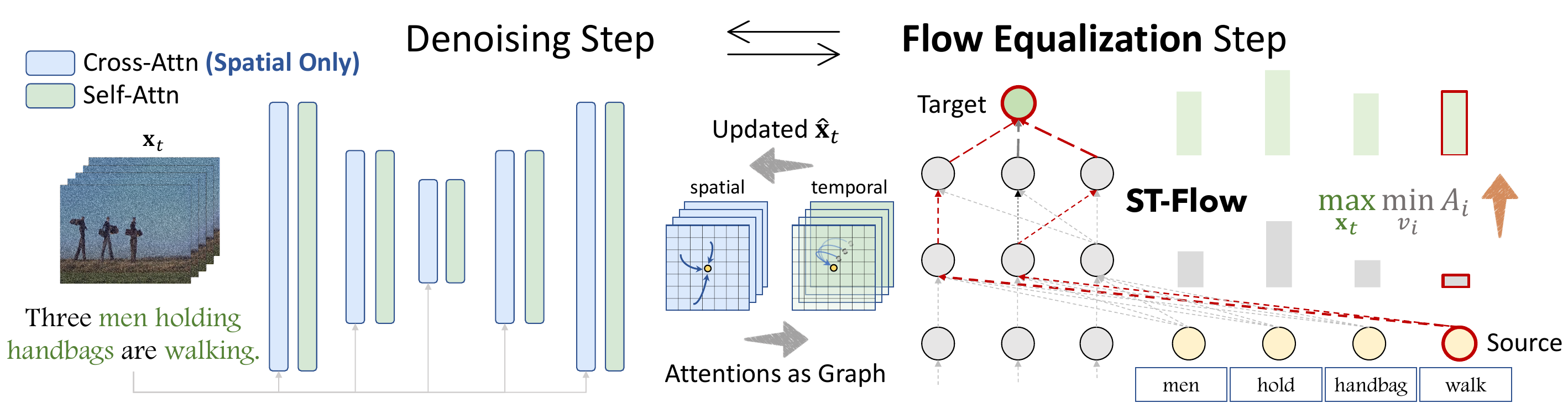}
    \caption{Overall pipeline of our \textbf{Vico}. Before each denoising step, Vico extracts attention maps from each layer to build a spatiotemporal graph. We calculate the attribution scores as max-flow in the graph and adjust the noisy latent code to balance this flows. }
    \label{fig:pipeline}
\end{figure}
\subsection{Overall Pipeline and Optimization}
\label{sec:prob&opt}
Our objective is to generate a video from an input prompt $P$, with $K$ text tokens  of interest $\mathcal{V}=\{v_1,\dots,v_K\}$. We aim to ensure that each token fairly contributes to the final video. This process is detailed in Figure~\ref{fig:pipeline}.

\noindent\textbf{Objectives.} To achieve this, we define an attribution function $A_i = A(v_i) \in \mathbb{R}$ for each token $v_i$, quantifying its impact on the video. We optimize the attribution scores to ensure fairness:
\begin{align}
    \max_{\mathbf{x}_t} \mathcal{L}_{\text{fair}}(A_1, \dots, A_K) = \max_{\mathbf{x}_t} \min_{v_i}\{A_1, \dots, A_K\};
    \label{eq:opt}
\end{align}
Here, $\mathcal{L}_{\text{fair}}=\min_{v_i}\{A_1, \dots, A_K\}$ serves as the fairness function, focusing on the least represented token. By updating the noisy latent $\mathbf{x}_t$ to maximize $\mathcal{L}_{\text{fair}}$, we ensure equal contributions across all tokens. Specifically, we estimate $A_i$ as flow in attention graph, which will be discussed later.

\noindent\textbf{Optimization.} To implement Eq~\ref{eq:opt}, we perform test-time optimization. Before each denoising step, we first feed $\mathbf{x}_t$ into the model, extract the $A_i$, and update $\mathbf{x}_t$ through gradient ascent: $\hat{\mathbf{x}}_t \leftarrow \mathbf{x}_t + \eta\nabla_{\mathbf{x}_t}\mathcal{L}_{\text{fair}}(A_1, \dots, A_K)$. $\eta$ is the step size. Then, $\hat{\mathbf{x}}_t$ is going through a denoising step to get $\mathbf{x}_{t-1}$ according to Eq~\ref{eq:update}. We repeat these steps until the video is generated.

\subsection{Attention Flow Across Space and Time}
\label{sec:ST-flow}
With above formulation, our focus is to develop an efficient and precise attribution $A_i$. Recognizing issues with cross-attention, we instead calculate $A_i$ as the flow through the entire attention graph.

\noindent\textbf{Flawed Cross-Attention in Text-to-Video Models.} Cross-attention score has been instrumental in attributing~\cite{tang-etal-2023-daam} and controlling layout and concept composition in text-to-image models~\cite{hertz2022prompt,10.1145/3592116,rassin2024linguistic}. However, applying it to T2V diffusion model introduces new problem. 

This problem arises because T2V models typically employ cross-attention on spatial tokens only~\cite{wang2023modelscope,chen2023videocrafter1,wang2023lavie}. It treats the video as a sequence of independent images, and temporal self-attention mixes tokens across different frames. Consequently, this separation hinders cross-attention's ability to capture video dynamics, making it challenging to manage actions across frames.

\begin{figure}[tb]
    \centering
    \includegraphics[width=\linewidth]{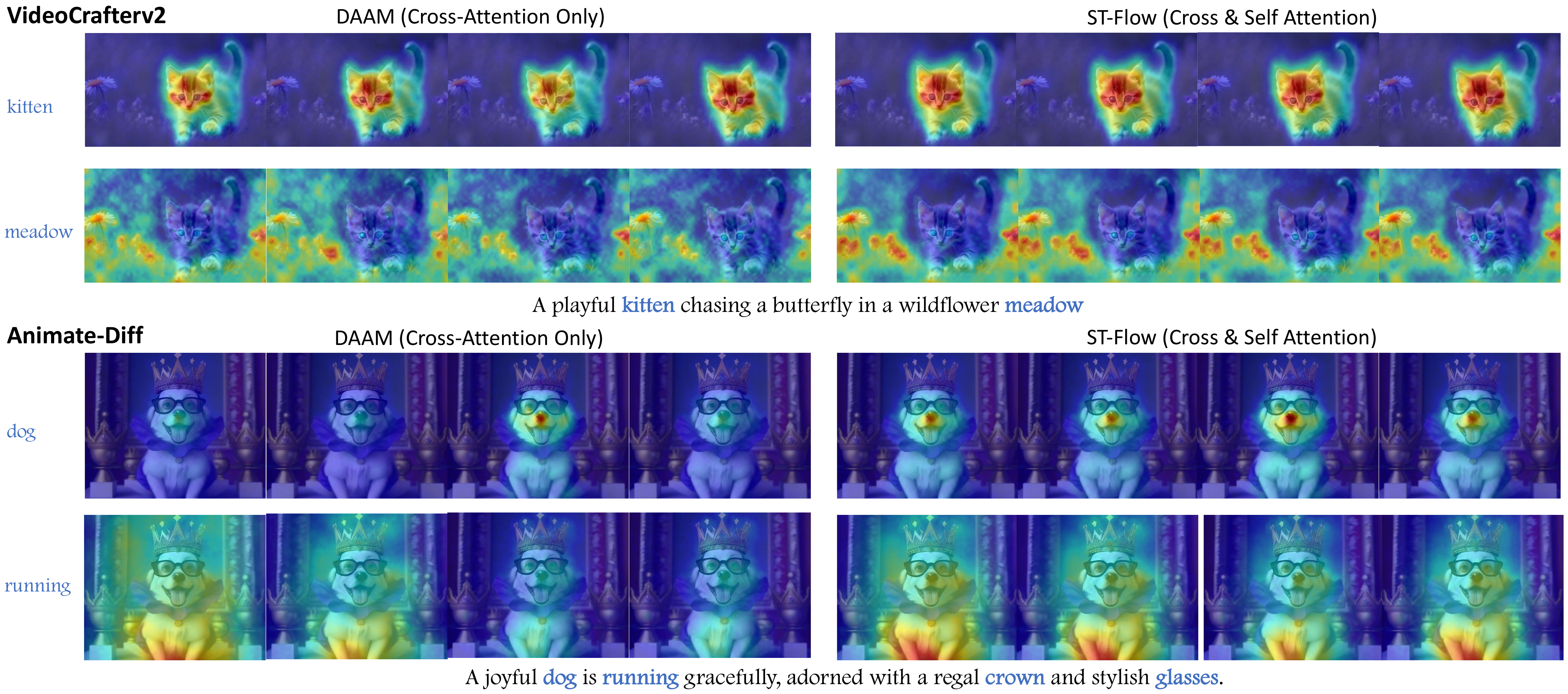}
    \caption{Attribution heatmap comparison between DAAM and ST-Flow.}
    \label{fig:attnvis}
\end{figure}

For example, applying the cross-attention-based DAAM attribution~\cite{tang-etal-2023-daam} on VideoCrafterv2 reveals significant issues in visualization. As shown in Figure~\ref{fig:attnvis} (Left), cross-attention leads to a flickering pattern in the attention maps, failing to consistently highlight the target object across frames.

Recognizing these limitations, we propose a new measurement termed \emph{Spatial-Temporal Flow (ST-Flow)}, which estimates the influence throughout the entire spatiotemporal attention graph in the video diffusion model. As seen in Figure~\ref{fig:attnvis} (Right), ST-Flow gets heatmap with improved consistency.

\noindent\textbf{Attention as a Graph Over Space and Time.} In our approach, we conceptualize the stacked attention layers as a directed graph $G=(V,E)$, where nodes represent tokens and edges weighted by the influence between tokens. A 4-layer example is illustrated in Figure~\ref{fig:pipeline} (Right). 

Its adjacency matrix is built using attention weights and skip connections~\cite{DBLP:conf/acl/AbnarZ20}. Suppose $w^{att}_{i,j}$ is the $i$-th row $j$-th column element of attention matrix averaged across heads.
For self-attention, the edge weight $e_{i,j}$ between any two tokens, $i$ and $j$, is $e_{i,j} = w^{att}_{i,j}+1$ if $i= j$, indicating a skip-connection, and $e_{i,j} = w^{att}_{i,j}$ if $i\neq j$. In the case of cross-attention, edge $e_{i,j} = w^{att}_{i,j}$ connects text to video, and $e_{i,i} = 1$ for connections within video tokens due to skip connections. Given that connections only exist from one layer to the next, the resulting matrix exhibits block-wise sparsity pattern. This is expressed as $\bm W = 
\begin{bmatrix}\bm 0 & E_{t,1}  & \bm 0 & \dots & E_{t,l} \\
\bm 0 & \bm 0 &E_{2} & \dots & \bm 0 \\
\bm 0 & \bm 0 &\bm 0 & \dots & \vdots \\
\bm 0 & \bm 0 &\bm 0 & \dots & E_{l} \\ \end{bmatrix}$. Here, $E_{i}$ denotes the edge weights within video, and $E_{t,i}$ indicates the influence from text to video at each cross-attention layer.

\textbf{Attribution as Flow on Graph.} Given graph $G$, we compute the attribution $A_i$ by analyzing all paths from a text token $v_i$ to video tokens at the output layer. As such, we formulate it as a \emph{max-flow problem} with capacity matrix $\bm W$. To facilitate this, we add an auxiliary target node $\bm v_{t}$ to $G$, connecting it to all output video tokens with inflow edges $e_{\bm v_{t}^{+}} = 1$\footnote{The maximum inflow is 1 for each node due to softmax normalization in the attention.}. We treat each text token $v_i$ as the source, and $\bm v_{t}$ as the sink. The max-flow from source to sink quantifies the influence of $v_i$, termed \emph{ST-Flow}.

\noindent\textbf{Definition 1} (ST-Flow). \textit{In attention graph $G$ with capacity matrix $\bm W$, a input token $v_i$ as source and sink node $\bm v_t$, the attribution value of $A_i = |f|^*$ is computed as the maximum flow from $v_i$ to $\bm v_t$.}

Our ST-Flow can be considered as an extension of Attention Flow~\cite{DBLP:conf/acl/AbnarZ20}, incorporating all attention layers in diffusion model. It is proved to be a kind of Shapley Value~\cite{DBLP:conf/acl/EthayarajhJ20}, which is an ideal contribution allocation in game theory~\cite{shapley1953value,myerson1977graphs,young1985monotonic} and interpretable AI model~\cite{lundberg2017unified,sundararajan2017axiomatic}.

\noindent\textbf{Exact ST-Flow Computation is infeasible.} While theoretically possible, calculating the ST-Flow in T2V diffusion models faces practical issues that render it infeasible:
\begin{itemize}[itemsep=0.1cm,parsep=0.1cm,leftmargin=*]
    \item \textbf{Non-Differentiable.} The max-flow algorithm, by its nature, is non-differentiable. This is a problem when we do gradient-based optimization in Eq~\ref{eq:opt}.
    \item \textbf{Efficiency Issue.} Solving max-flow for each input token is slow. Even with the Dinic's algorithm~\cite{dinic1970algorithm}~\footnote{Given that the attentions has more edge than tokens, Dinic is best choice in theory. However, our implementation shows that max-flow on each token takes $\sim$8s.}, the time complexity is $O(K|V|^2 |E|)$ for large attention graphs extracted from video diffusion model.
\end{itemize}
Despite these obstacles, in Sec~\ref{sec:approaximation}, we derive a min-max approximation to circumvent these issues.

\subsection{Differentiable ST-Flow with Min-Max Path Flow}
\label{sec:approaximation}
As discussed above, exact computation of ST-Flow is challenging.
Instead of directly estimating the ST-Flow, we approach this by focusing on approximating its lower bound, which is computationally feasible. This is made possible, since any sub-graph has max-flow smaller than that of full graph.

\textbf{Theorem 1} (Sub-Graph Flow)\footnote{Proof in Appendix~\ref{proof:subflow}}. \textit{For any sub-graph $g$ of a graph $G$, $g \subseteq G$, the maximum flow $f^*_g$ in $g$ is less than or equal to the maximum flow $f^*_G$ in $G$, $|f^*_g| \leq |f^*_G|$.}

Based on this theorem, we need not compute the ST-Flow directly. Instead, we sample multiple subgraphs $g$ from $G$, calculate the maximum flow for each, and take the highest value among these:
    {\small\begin{align}
    |f_G| \geq A_i = \max_{\forall g \subseteq G} |f_g|;
\end{align}}
This approach allows for a more efficient calculation by focusing on a manageable number of subgraphs, solving the max-flow for each, and identifying the maximum flow.

In this work, we focus on the simplest type of subgraph in graph $G$: a path from a $v_i$ to target $\bm v_t$. We efficiently approximate the ST-Flow by computing the \emph{max path flow} for each path. We propose two min-max strategies to achieve this: 
\begin{itemize}[itemsep=0.1cm,parsep=0.1cm,leftmargin=*]
    \item \textbf{Hard Flow Strategy.} For each text token $v$, we sample all paths $v_i$ to $v_t$. The max-flow on each path is calculated as the minimum edge capacity along the path, $|f| = \min_{j} e_j$. And the best approximated $A_i = \max |f|$ is the maximum of these minimums across all paths. 
    \item \textbf{Soft Flow Strategy.} Instead of get the hard min-max flow, we use \emph{soft-min} and \emph{soft-max} operations using the log-sum-exp trick. This approach provides a smoother approximation of flow values, which can be especially useful in our gradients-based optimization. The soft-min/max is computed as below, with $\tau$ as a temperature
    {\small
    \begin{align}
        \text{softmax}(e_1, e_2,\dots;\tau) = \tau \log\Big(\sum_{j} \exp{(\frac{e_j}{\tau})}\Big);\\
        \text{softmin}(e_1, e_2,\dots;\tau) = - \text{softmax}(-e_1, -e_2,\dots;\tau),
    \end{align}}
\end{itemize}

\noindent\textbf{Vectorized Path Flow Computation.} While depth-first and breadth-first searches can identify all paths for above min-max optimization, these methods are slow and cannot be parallelized.
Instead, we define a special operation called \emph{min-max multiplication} on the capacity matrix to calculate the maximum flow for each path in a vectorized manner.

\noindent\textbf{Definition 2} (Min-max Multiplication). \textit{Given two matrices $A\in \mathbb{R}^{m\times k}$ and $B\in \mathbb{R}^{k\times n}$, min-max multiplication $C=A \odot B \in \mathbb{R}^{m\times n}$ is defined where each element $C_{i,j} = \max_r (\min(A_{i,r}, B_{r,j}))$.}

This operation computes the minimum value across all $r$ for the $i$-th row of $A$ and the $j$-th column of $B$, and $\max_{r}$ selects the maximum of these minimum values for each $C_{i,j}$. 
We call it a \emph{multiplication} because it resembles matrix multiplication but replaces element-wise multiplication with a minimum operation and summation with maximization. 

A very good property is that, the min-max multiplication of capacity matrix $\bm W^{k} = \bm W^{k-1} \odot \bm W$ can be interpreted as the max path flow for all $k$-hop paths.

\noindent\textbf{Proposition 1} (Max Path Flow using Min-max Multiplication)\footnote{Proof in Appendix~\ref{proof: maxpathflow}}. \textit{For min-max power of capacity $\bm W^{k} = \bm W^{k-1} \odot \bm W$, element $\bm W^{k}_{i,j}$ equals the max path flow for all $k$-hop path from $v_i$ to $v_j$.}

For attention graph that current layer's node is only connect to the next layer, all path from text token to output video token has exactly the length of $l$. In this way, what we do is just to extract the attention graph $G$, do $l$ times Min-max Multiplication on its flow matrix, and we consider the value as a approximation of ST-Flow. A tine complexity analysis is prepared in Appendix~\ref{sec:speed}.

In this way, we get all pieces to build \textbf{Vico}. We first compute attribution using the approximated ST-Flow, then using Eq~\ref{eq:opt} to update the latent to equalize such flow.

\section{Experiments}
In our experiments section, we evaluate \textbf{Vico} through a series of tests. We start by assessing its performance on generating videos from compositional text prompts. Next, we demonstrate ST-Flow accurately attributes token influence through video segmentation and human study. We also conduct an ablation study to validate our key designs. More application results are provide in Appendix~\ref{sec:edit}.

\subsection{Experiment Setup}

\noindent\textbf{Baselines.} We build our method on several open-sourced video diffusion model, including VideoCrafterv2~\cite{chen2024videocrafter2}, AnimateDiff~\cite{guo2023animatediff} and Zeroscopev2~\footnote{\url{https://huggingface.co/cerspense/zeroscope_v2_576w}}. Since no current video diffusion models specifically focus on generation with compositional conditions, we re-implement several methods designed for text-to-image diffusion models and compare with them. These methods include:
\begin{itemize}
[itemsep=0.1cm,parsep=0.1cm,leftmargin=*]
    \item \textit{Original Model}. We directly ask the original base model to produce video based on prompts.
    \item \textit{Token Re-weight}. We use the \texttt{compel}~\footnote{\url{https://github.com/damian0815/compel}} package to directly up-lift the weight of specific concept token, with a fixed weight of 1.5.
    \item \textit{Compositional Diffusion}~\cite{liu2022compositional}. This method directly make multiple noise predictions on different text, and sum the noise prediction as the compositional direction for latent update. In our paper, given a prompt, we first split into short phrases. For example ``\texttt{a dog and a cat}'' is splitted into ``\texttt{a dog}'' and ``\texttt{a cat}'', make individual denoising, and added up.
    \item \textit{Attend-and-Excite}~\cite{10.1145/3592116}. A\&E refines the noisy latents to excite cross-attention units to attend to all subject tokens in the text prompt.
\end{itemize}

\begin{table}[]
\scriptsize
\renewcommand{\arraystretch}{0.85}
    \caption{Quantitative results for different methods on compositional text-to-video generation.}
    \label{tab:quant_comp}
    \vspace{-2mm}
    \centering
    \begin{tabular}{l|c|c|c|c}
    \toprule
        Name & Spatial Relation$\uparrow$ &  Multiple Object$\uparrow$ & Motion Composition$\uparrow$ & Overall Consistency$\uparrow$\\
            \midrule
        AnimateDiff~\cite{guo2023animatediff}  &24.80\%& 33.44\% & 33.90\% & 27.75\%\\
        +Compositional Diffusion~\cite{liu2022compositional} & 19.43\%  & 7.27\% & 23.58\% & 24.07\% \\
        +Attend-and-Excite~\cite{10.1145/3592116} &20.88\% & 31.25\% & 34.78\% & 28.05\% \\
        +Token-Reweight &28.11\% & 36.89\%&37.45\% & 26.77\% \\ 
        \midrule
         \rowcolor{gray!5}+\texttt{Vico}~(\textit{hard}) & 24.22\% & 29.95\%& 37.23\% &28.85\% \\
        \rowcolor{gray!10}+\texttt{Vico}~(\textit{soft}) &\textbf{31.47}\%& \textbf{37.20}\% & \textbf{37.95}\% &\textbf{28.89}\%\\
        \midrule\midrule
        ZeroScopev2 & 59.52\% & 52.52\% & 45.51\% & 25.83\% \\
         +Compositional Diffusion~\cite{liu2022compositional} & 31.77\%& 8.23\%&33.13\% & 23.02\% \\
        +Token-Reweight & 57.48\% & 50.00\% &40.42\% &25.74\%\\
        +Attend-and-Excite~\cite{10.1145/3592116} & 59.02\%& 62.27\% & 45.82\% & 25.84\%\\
        \midrule
        \rowcolor{gray!5}+\texttt{Vico}~(\textit{hard}) & \textbf{63.60}\% & 63.34\% &\textbf{46.32}\% & 24.89\% \\
        \rowcolor{gray!10}+\texttt{Vico}~(\textit{soft}) & 62.28\% & \textbf{69.05}\% & 45.31\% &\textbf{26.15}\%\\
        
        \midrule\midrule
        VideoCrafterv2~\cite{chen2023videocrafter1} & 35.86\% &  40.66\%& 43.82\% & 28.06\%\\
        +Compositional Diffusion~\cite{liu2022compositional} & 23.61\% & 10.59\%&35.49\% & 24.49\% \\
        +Token-Reweight & 46.08\% & 49.16\% & 44.33\% & 28.29\%\\
        +Attend-and-Excite~\cite{10.1145/3592116} & 48.11\% & 66.62\%  & 43.48\% & 28.33\%\\
        \midrule
        \rowcolor{gray!5}+\texttt{Vico}~(\textit{hard}) & 49.85\% & 67.84\% & 44.46\% & 28.41\%\\
        \rowcolor{gray!10}+\texttt{Vico}~(\textit{soft}) & \textbf{50.40}\% &  \textbf{73.55}\% & \textbf{44.98}\% & \textbf{28.52}\%\\
         \bottomrule
    \end{tabular}
    \vspace{-3mm}
\end{table}

\noindent\textbf{Evaluation and Metrics.} We evaluate compositional generation using VBench~\cite{huang2023vbench}. Specifically, we focus on evaluating compositional quality in terms of \emph{Spatial Relation}, \emph{Multiple Object Composition}. For both metrics, the model processes text containing multiple concepts, generates a video. Then a caption model verifies the accuracy of the concept representations within the generated video.

Additionally, we design a new metric, \emph{Motion Composition}. This metric evaluates the generated video based on the presence and accuracy of multiple objects performing different motions. We collect 70 prompts of the form "\texttt{obj}$_1$ is \texttt{motion}$_1$ and \texttt{obj}$_2$ is \texttt{motion}$_2$". Using GRiT~\cite{wu2022grit}, we generate dense captions on video for each object and verify if each \texttt{(object, motion)} pair appears in the captions. The score is computed as $\frac{\sum_{1,2} (\mathbb{I}(\texttt{obj}_i) + \mathbb{I}(\texttt{obj}_i, \texttt{motion}_i)) }{4}$. Here, $\mathbb{I}(x)$ is an indicator function that returns 1 if $x$ is present in the generated captions, and 0 otherwise.

The overall video quality is measured using ViCLIP~\cite{wang2023internvid} to compute a score based on text and video alignment, denoted as \emph{Overall Consistency}. 

\noindent\textbf{Implementation Details.} We use the implementations on \texttt{diffusers} for video generation. All videos are generated by a A6000 GPU. We sample videos from Zeroscopev2 and VideoCrafterv2 using 50-step DPM-Solver++~\cite{lu2022dpm}. AnimateDiff is sampled with 50-step DDIM~\cite{song2020denoising}. We optimize the latent at each sampling steps, and update the latent with Adam~\cite{kingma2014adam} optimizer at the learning rate of $1e-5$. We test both the soft and hard-min/max versions of Vico, setting the temperature $\tau = 0.01$ for the soft version. The NLTK package identify all nouns and verbs for equalization.

\subsection{Compositional Video Generation}
\noindent\textbf{Quantitative Results.} In Table~\ref{tab:quant_comp}, we present the scores achieved by \textbf{Vico} compared to other methods across various base models on compositional text-to-video generation. Vico consistently surpasses all baselines on every metric. Notably, our ST-flow based method surpasses cross-attention based techniques like Attend\&Excite, thanks to its ability to incorporating influences across full attention graph. Additionally, the soft min-max version of Vico generally achieves better fidelity than the hard version, as it is better suited for gradient optimization.

Specifically, Vico demonstrates its most significant improvements in multi-subject generation tasks. For instance, on VideoCrafterv2, it shows a remarkable increase, improving scores from $40.66\% \to 73.55\%$. This suggests that our attention mechanism in T2V is more adept at managing object arrangement.  In contrast, compositional diffusion models~\cite{liu2022compositional} often fail, as they assume conditions to be independent, which is problematic for realistic condition compositions.

\begin{figure}
    \centering
    \includegraphics[width=\textwidth]{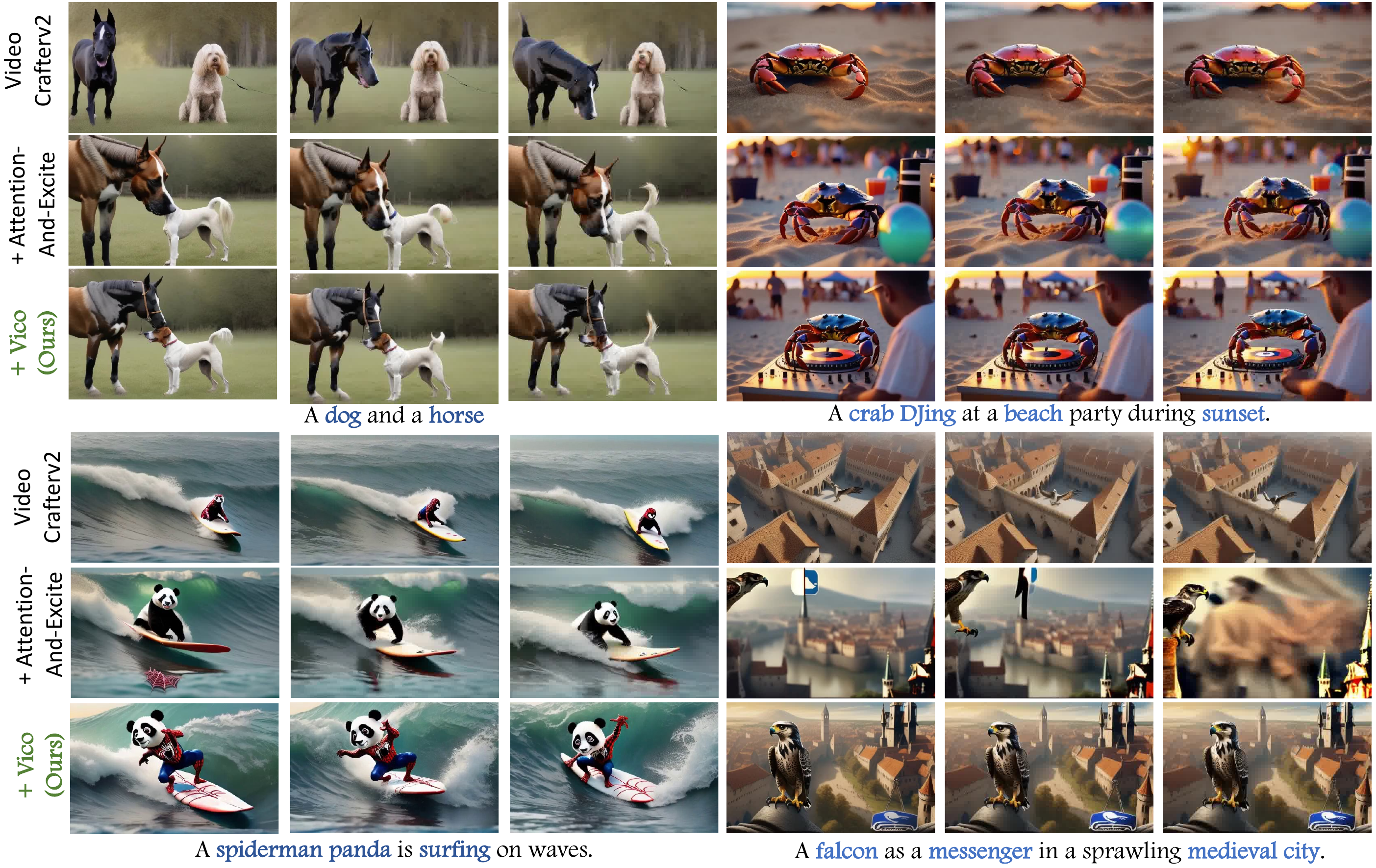}
    \caption{Qualitative comparison of the videos generated by VideoCrafterv2 baseline, Attribute\&Excite and our \textbf{Vico} with compositional textual descriptions.}
    \label{fig:quality}
\end{figure}

\noindent\textbf{Qualitative Results.} We compare the videos generated by different methods in Figure~\ref{fig:quality}. Attend\&Excite receive slightly improvements, but still mixes semantics of different subject. For example, on the ``\texttt{a dog and a horse}'' example (Top Left), both Attend\&Excite and the baseline incorrectly combine a dog's face with a horse's body. Vico addresses this issue by ensuring each token contributes equally, effectively separating their relationships.

Additionally, cross-attention often leads to temporal inconsistencies in the modified videos. For instance, in the ``\texttt{spider panda}'' case (Bottom Left), Attend\&Excite  initially displays a Spider-Man logo but it disappears abruptly in subsequent frames. In contrast, Vico captures dynamics across both spatial and temporal attention, leading to better results. More results is in Appendix~\ref{sec:edit} and~\ref{sec:more}.

\subsection{Attribution on Video Diffusion Model}
In this section, we aim to demonstrate that our ST-Flow~(hard) provides a more accurate measure of token contribution compared to other attention-based indicators. 

\noindent\textbf{Objective Evaluation: Zero-shot Video Segmentation.} We test several attribution methods using the VideoCrafterv2 model for zero-shot video segmentation on the Ref-DAVIS2017~\cite{khoreva2019video} dataset. To create these maps, we first performed a 25-step DDIM inversion~\cite{mokady2023null} to extract noise patterns, followed by sampling to generate the attribution maps. We specifically use maps from from \emph{end of text}~([EOT]) token~\cite{li2024get} for segmentation. We used the mean value of the map as a threshold for binary segmentation. We compare with cross-attention~\cite{tang-etal-2023-daam} and Attention Rollout~\cite{DBLP:conf/acl/AbnarZ20}. The more accurate the segmentation is, the attribution is more reasonable for human.

Results are presented in Table~\ref{tab:performance_ref_david2017}. Our method outperform others, providing the highest segmentation metrics in zero-shot setting. As visualized in Figure~\ref{fig:seg}, cross-attention maps show inconsistent highlighting and flickering. Attention Rollout also conciders the full attention graph, but overly smoothed weights, resulting in less precise object focus.

\noindent\textbf{Subjective Evaluation: User Study.} Besides, segmentation-based validation, we conducted a subjective user study to evaluate the quality of attribution maps generated by various methods. 20 participants rated maps from three different approaches across ten video clips. The evaluation focused on \emph{Temporal Consistency}, assessing the presence of flickering, and \emph{Reasonability}, determining alignment with human interpretations. Ratings ranged from 1 to 5, with 5 as the highest. As summarized in Table~\ref{tab:attribution_methods}, Our ST-Flow outperforms other attention-based explanation, achieving the highest scores in both Temporal Consistency (4.19) and Reasonability (3.78).

\begin{table}[t]
\centering
\begin{minipage}[c]{0.48\linewidth}
\renewcommand{\arraystretch}{1.03}
\setlength{\tabcolsep}{4pt}
\centering
\scriptsize
\begin{tabular}{l|c | c}
\toprule
    Attribution Method & Temporal Consistency$\uparrow$ & Reasonability$\uparrow$ \\
    \midrule
Cross-Attention~\cite{tang-etal-2023-daam} & 2.56 & 2.84  \\
    Attention Rollout~\cite{DBLP:conf/acl/AbnarZ20} & 3.88 & 3.45 \\
    \midrule
     ST-Flow (Ours) &\textbf{4.19} & \textbf{3.78} \\
    \bottomrule
\end{tabular}
\caption{User study on attribution method.}
\label{tab:attribution_methods}
\vspace{-1mm}
\renewcommand{\arraystretch}{1.0}
\setlength{\tabcolsep}{3pt}
\scriptsize
\centering
\begin{tabular}{l|ccc}
\toprule
   \multirow{2}{*}{Method} & \multicolumn{3}{c}{Ref-DAVID2017}  \\
   \cmidrule{2-4}
     & $\mathcal{J}$\&$\mathcal{F}\uparrow$ &$\mathcal{J}\uparrow$ & $\mathcal{F}\uparrow$\\
     \hline
    \multicolumn{4}{c}{\textcolor{gray}{Supervised Trained}} \\
    \hline
    \textcolor{gray}{ReferFormer-B}~\cite{wu2022language} & \textcolor{gray}{61.1} & \textcolor{gray}{58.1} & \textcolor{gray}{64.1} \\
    \textcolor{gray}{OnlineRefer-B}~\cite{wu2023onlinerefer} & \textcolor{gray}{62.4} & \textcolor{gray}{59.1} & \textcolor{gray}{65.6} \\
    \hline
     \multicolumn{4}{c}{Zero-Shot} \\
    \hline
    Cross-Attention~\cite{tang-etal-2023-daam} mean & 32.1 & 29.8& 34.7 \\
    Attention Rollout~\cite{DBLP:conf/acl/AbnarZ20} mean & 38.0 & 33.3& 40.0 \\
    ST-Flow (Ours) mean & \textbf{38.2} & \textbf{33.5}& \textbf{40.3}\\
    \bottomrule
\end{tabular}
\caption{Performance on Ref-DAVID2017.}
\label{tab:performance_ref_david2017}
\end{minipage}
\hfill
\begin{minipage}[c]{0.48\linewidth}
\vspace{-1mm}
\setlength{\tabcolsep}{2pt}
\scriptsize
\begin{tabular}{cc|c|c}
    \toprule
        Min Loss & ST-Flow~\textit{(soft)} & Multiple Object$\uparrow$ & Overall Consistency$\uparrow$ \\
         \midrule
            \textcolor{red}{\xmark}   & \textcolor{red}{\xmark} & 57.86\% & 28.03\%\\
            
            \textcolor{applegreen}{\cmark} & \textcolor{red}{\xmark} & 63.62\% & 28.24\%\\
            \textcolor{red}{\xmark}  & \textcolor{applegreen}{\cmark} & 69.75\% & 28.12\% \\
\textcolor{applegreen}{\cmark}   & \textcolor{applegreen}{\cmark} &\textbf{73.55}\% & \textbf{28.52}\% \\
        \bottomrule
    \end{tabular}
    \caption{Ablation study on Vico.}
    \label{tab:ablate}
    \vspace{-4mm}
\begin{center}
    \includegraphics[width=1.02\textwidth]{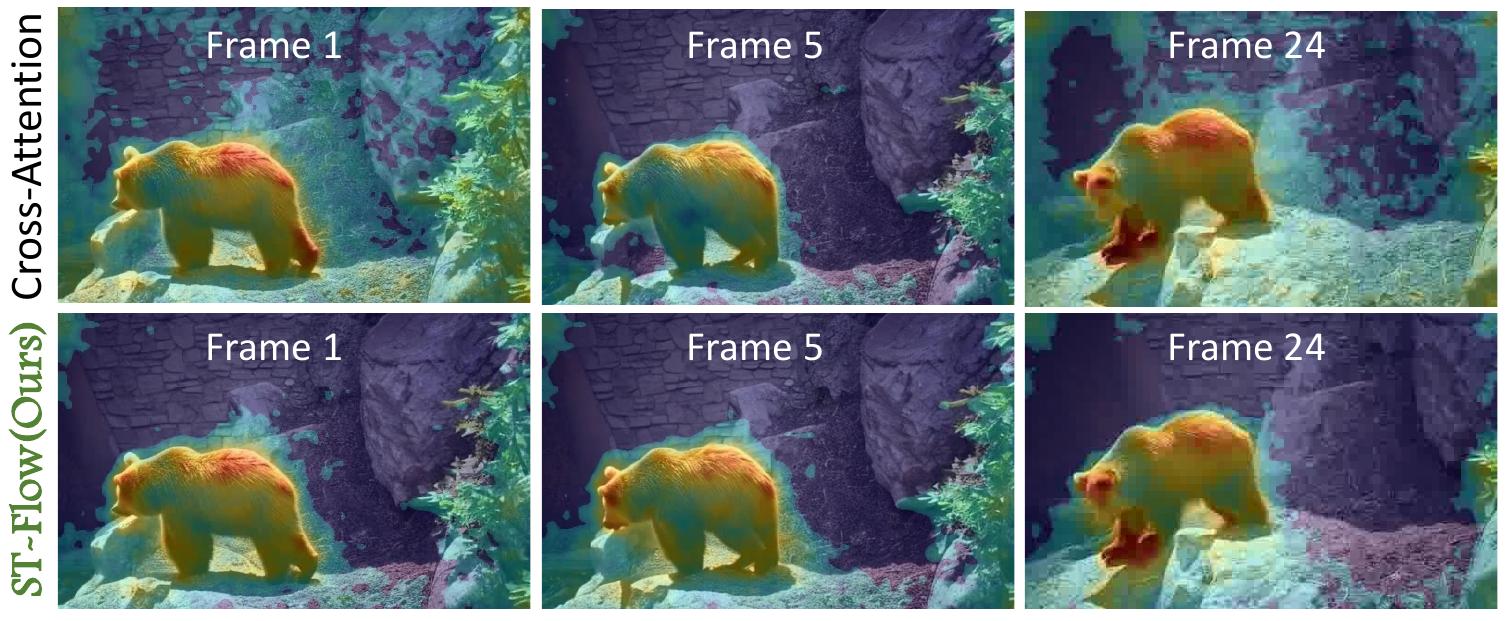} 
  \end{center}
  \vspace{-2mm}
  \caption{Segmentation results comparison.}
  \label{fig:seg}
\end{minipage}
\vspace{-6mm}
\end{table}
\subsection{Ablation Study}
In this section, we ablate our two key designs: the loss function and the proposed ST-Flow. We do experiments on VideoCrafterv2 and compare the performance.

\noindent\textbf{Loss Function.} We modify the loss function used, from using the ``min'' as a fairness indicator (as described in Sec~\ref{sec:prob&opt}) to a variance loss, defined as $\mathcal{L}_{\text{fair}} = -\sum_{i}(A_i-\Bar{A})^2$. It minimizes the differences between each $A_i$ and the average attribution value $\Bar{A}$, making it fair. The results are shown in Table~\ref{tab:ablate}, row 3 and 4. We notice while the variance loss ensures uniformity across all tokens, it overly restricts them, often degrading video quality. Conversely, our min-loss focuses on the least represented token, enhancing object composition accuracy without significantly affecting overall quality.

\noindent\textbf{ST-Flow $v.s.$ Cross-Attention.} A major contribution of our work is the development of ST-Flow and its efficient computation. We compare it against a variant using cross-attention as attribution. In this method, cross-attention maps are extracted and a mean score is computed for each token as $A_i$. As demostrated in Table~\ref{tab:ablate}, row 2 and 4, using ST-Flow (soft) largely outperforms cross-attention. We also provide the running speed analysis in Appendix~\ref{sec:speed}, confirming the efficiency of our approach.

\section{Conclusion}
In this paper, we present \textbf{Vico}, a framework designed for compositional video generation. Vico starts by analyzing how input tokens influence the generated video. It then adjusts the model to ensure that no single concept dominates. To implement Vico practically, we calculate each text token's contribution to the video token using max flow. This computation is made feasible by approximating the subgraph flow with a vectorized implementation. We have applied our method across various diffusion-based video models, which has enhanced both the visual fidelity and semantic accuracy of the generated videos.

{
    \small
    \bibliographystyle{plain}
    \bibliography{neurips_2024}
}


\appendix



\section{Proof of Theorem 1: Sub-graph Flow}
\label{proof:subflow}
In a network \( G = (V, E) \) with a capacity function \( c: E \rightarrow \mathbb{R}^+ \), and a subgraph \( g \) of \( G \), the maximum flow \( f_g \) in \( g \) is less than or equal to the maximum flow \( f_G \) in \( G \).

\subsection*{Proof}
\begin{enumerate}
    \item \textbf{Definition of a Subgraph:} A subgraph \( g \) of \( G \) can be defined as \( g = (V', E') \) where \( V' \subseteq V \) and \( E' \subseteq E \). All capacities in \( g \) are inherited from \( G \), i.e., \( c'(e) = c(e) \) for all \( e \in E' \).

    \item \textbf{Flow Conservation:} Both \( G \) and \( g \) must satisfy the flow conservation law at all intermediate nodes. That is, the sum of the flow entering any node must equal the sum of the flow exiting that node, except for the source (where flow is generated) and the sink (where flow is absorbed).

    \item \textbf{Reduced Set of Paths:} Since \( E' \subseteq E \), every path through \( g \) is also a path through \( G \), but not every path through \( G \) is necessarily a path through \( g \). This reduction in the number of paths (or edges) in \( g \) implies that some routes available for flow in \( G \) are not available in \( g \).

    \item \textbf{Capacity Limitations:} For any edge \( e \) in \( E' \), the capacity in \( g \) (i.e., \( c'(e) \)) equals the capacity in \( G \) (i.e., \( c(e) \)). Therefore, no edge in \( g \) can support more flow than it can in \( G \). Additionally, since some edges might be missing in \( g \), the overall capacity of pathways from the source to the sink might be less in \( g \) than in \( G \).

    \item \textbf{Maximum Flow Reduction:} Given the reduction in paths and capacities, any flow that is feasible in \( g \) is also feasible in \( G \), but not vice versa. Hence, the maximum flow \( f_g \) that can be pushed from the source to the sink in \( g \) must be less than or equal to the maximum flow \( f_G \) that can be pushed in \( G \).
\end{enumerate}

\textbf{Conclusion:} From these points, it follows directly that the maximum flow in a subgraph \( g \) cannot exceed the maximum flow in the original graph \( G \). This proves that \( f_g \leq f_G \).


\section{Proof of Proposition 1: Max Path Flow using Min-max Multiplication}
\label{proof: maxpathflow}
\textbf{Definitions and Proposition:}
Let $\mathbf{W}$ be a capacity matrix of a graph where $\mathbf{W}_{i,j}$ is the capacity of the edge from vertex $i$ to vertex $j$. If there is no edge between $i$ and $j$, $\mathbf{W}_{i,j} = 0$ or some representation of non-connectivity. A $k$-hop path between two vertices $i$ and $j$ is a path that uses exactly $k$ edges.

\textbf{Proposition:} The $k$-th min-max power of $\mathbf{W}$, denoted $\mathbf{W}^k$, calculated as $\mathbf{W}^k = \mathbf{W}^{k-1} \odot \mathbf{W}$, has elements $\mathbf{W}^k_{i,j}$ that represent the maximum flow possible on any $k$-hop path from vertex $i$ to $j$.

\textbf{Min-max Multiplication:}
Given matrices $\mathbf{A}$ and $\mathbf{B}$, $\mathbf{C} = \mathbf{A} \odot \mathbf{B}$ is defined such that:
\[
\mathbf{C}_{i,j} = \max_r (\min(\mathbf{A}_{i,r}, \mathbf{B}_{r,j}))
\]

\textbf{Proof by Induction:}

\textbf{Base Case} ($k = 1$):
\begin{itemize}
    \item \textbf{Claim:} $\mathbf{W}^1_{i,j}$ represents the capacity of the edge from $i$ to $j$, which is the maximum flow on a 1-hop path.
    \item \textbf{Proof:} By definition, $\mathbf{W}^1 = \mathbf{W}$, and $\mathbf{W}^1_{i,j} = \mathbf{W}_{i,j}$, which directly corresponds to the edge capacity between $i$ and $j$. Hence, the base case holds.
\end{itemize}

\textbf{Inductive Step:}
\begin{itemize}
    \item \textbf{Assumption:} Assume that for $k-1$, $\mathbf{W}^{k-1}_{i,j}$ correctly represents the maximum flow on any $k-1$-hop path from $i$ to $j$.
    \item \textbf{To Prove:} $\mathbf{W}^k_{i,j}$ represents the maximum flow on any $k$-hop path from $i$ to $j$.
\end{itemize}

\textbf{Proof:}
From the definition of min-max multiplication,
\[
\mathbf{W}^k_{i,j} = \max_r (\min(\mathbf{W}^{k-1}_{i,r}, \mathbf{W}_{r,j}))
\]
\begin{itemize}
    \item $\mathbf{W}^{k-1}_{i,r}$ is the maximum flow from $i$ to $r$ using $k-1$ hops.
    \item $\mathbf{W}_{r,j}$ is the capacity of the edge from $r$ to $j$ (1-hop).
\end{itemize}
\textbf{Interpretation:} $\min(\mathbf{W}^{k-1}_{i,r}, \mathbf{W}_{r,j})$ finds the bottleneck flow for the path from $i$ to $j$ through $r$ using $k$ hops. The minimum function ensures the path's flow is constrained by its weakest segment.

\textbf{Maximization Step:} $\max_r$ over all possible intermediate vertices $r$ selects the path with the highest bottleneck value, thus ensuring the selected path is the most capable among all possible $k$-hop paths.

\textbf{Conclusion:}
The inductive step confirms that the flow represented by $\mathbf{W}^k_{i,j}$ is indeed the maximum possible flow across any $k$-hop path from $i$ to $j$. Hence, by induction, the proposition holds for all $k$.

\section{Related Work}
\label{sec:related}
\noindent\textbf{Video Diffusion Models.}  Video diffusion models generate video frames by gradually denoising a noisy latent space~\cite{DBLP:conf/nips/HoSGC0F22}. One of the main challenges with these models is their high computational complexity. Typically, the denoising process is performed in the latent space~\cite{zhou2022magicvideo,blattmann2023align,blattmann2023stable}. The architectural commonly adopt either a 3D-UNet~\cite{DBLP:conf/nips/HoSGC0F22,blattmann2023align,DBLP:journals/corr/abs-2210-02303,harvey2022flexible,wu2023tune} or diffusion transformer~\cite{DBLP:journals/corr/abs-2312-06662,DBLP:conf/iccv/PeeblesX23,DBLP:journals/corr/abs-2401-03048}. To enhance computational efficiency, these architectures often employ separate self-attention mechanisms for managing spatial and temporal tokens. Conventionally, training these models involves fine-tuning an image-based model for video data~\cite{wu2023tune,text2video-zero,guo2023animatediff}. This process includes adding a temporal module while striving to preserve the original visual quality.  

Despite their ability to generate photorealistic videos, these models frequently struggle with understanding the complex interactions between elements in a scene. This shortcoming can result in the generation of nonsensical videos when responding to complex prompts.

\noindent\textbf{Compositional Generation.} Current generative models often face challenges in creating data from a combination of conditions, with most developments primarily in the image domain.  Energy-based models~\cite{du2020compositional,du2023reduce,liu2023unsupervised}, for example, are mathematically inclined to be compositionally friendly, yet they require the conditions to be independent.
In practice, many image-based methods utilize cross-attention to effectively manage the composition of concepts~\cite{feng2023trainingfree,10.1145/3592116,wu2023harnessing,rassin2024linguistic}. However, when it comes to video, compositional generation introduces additional complexities. Some video-focused approaches concentrate specific form of composition, including object-motion composition~\cite{wei2023dreamvideo}, subject-composition~\cite{wang2024customvideo}, utilize explicit graphs to control content elements~\cite{pmlr-v139-bar21a}, or integrate multi-modal conditions~\cite{wang2024videocomposer}. Despite these efforts, a generic solution for accurately generating videos from text descriptions involving multiple concepts is still lacking. We present the first solution for compositional video generation using complex text prompts, an area that remains largely underexplored.

\noindent\textbf{Attribution Methods.} Attribution methods clarify how specific input features influence a model's decisions. gradient-based methods~\cite{sundararajan2017axiomatic, simonyan2013deep,selvaraju2017grad} identify influential image regions by back-propagating gradients to the input. Attention-based methods~\cite{chefer2021generic,DBLP:conf/acl/AbnarZ20} that utilize attention scores to emphasize important inputs. Ablation methods\cite{ramaswamy2020ablation,zeiler2014visualizing} modify data parts to assess their impact. Shapley values~\cite{NIPS2017_8a20a862} distribute the contribution of each feature based on cooperative game theory. In our paper, we extend existing techniques of attention flow to video diffusion models. We develop an efficient approximation to solve the max-flow problem.  This improvement helps us more accurately identify and balance the impact of each textual elements on synthesized video.

\section{Compositional Video Editing}
\label{sec:edit}
Our system, Vico, can be integrated into video editing workflows to accommodate text prompts that describe a composition of concepts.

\noindent\textbf{Setup.} We begin by performing a 50-step DDIM inversion on the input video. Following this, we generate a new video based on the given prompt.

\begin{figure}[H]
    \centering
    \includegraphics[width=\linewidth]{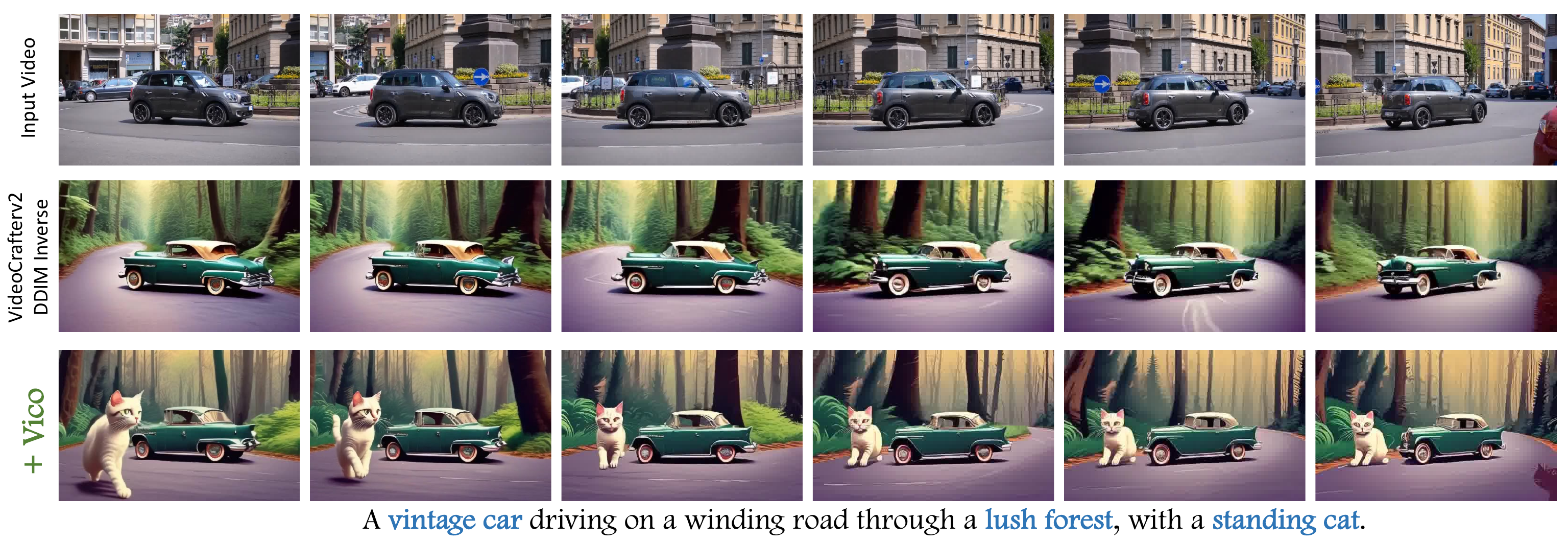}
    \caption{Video edit results with compositional prompts.}
    \label{fig:edit}
\end{figure}

\noindent\textbf{Results.} An example of this process is illustrated in Figure~\ref{fig:edit}. The original video demonstrates a strong bias towards a single presented object, making editing with a composition of concepts challenging. However, by applying Vico, we successfully enhance the video to accurately represent the intended compositional concepts.

\section{More Visualizations}
\label{sec:more}
\begin{figure}[H]
    \centering
    \includegraphics[width=\linewidth]{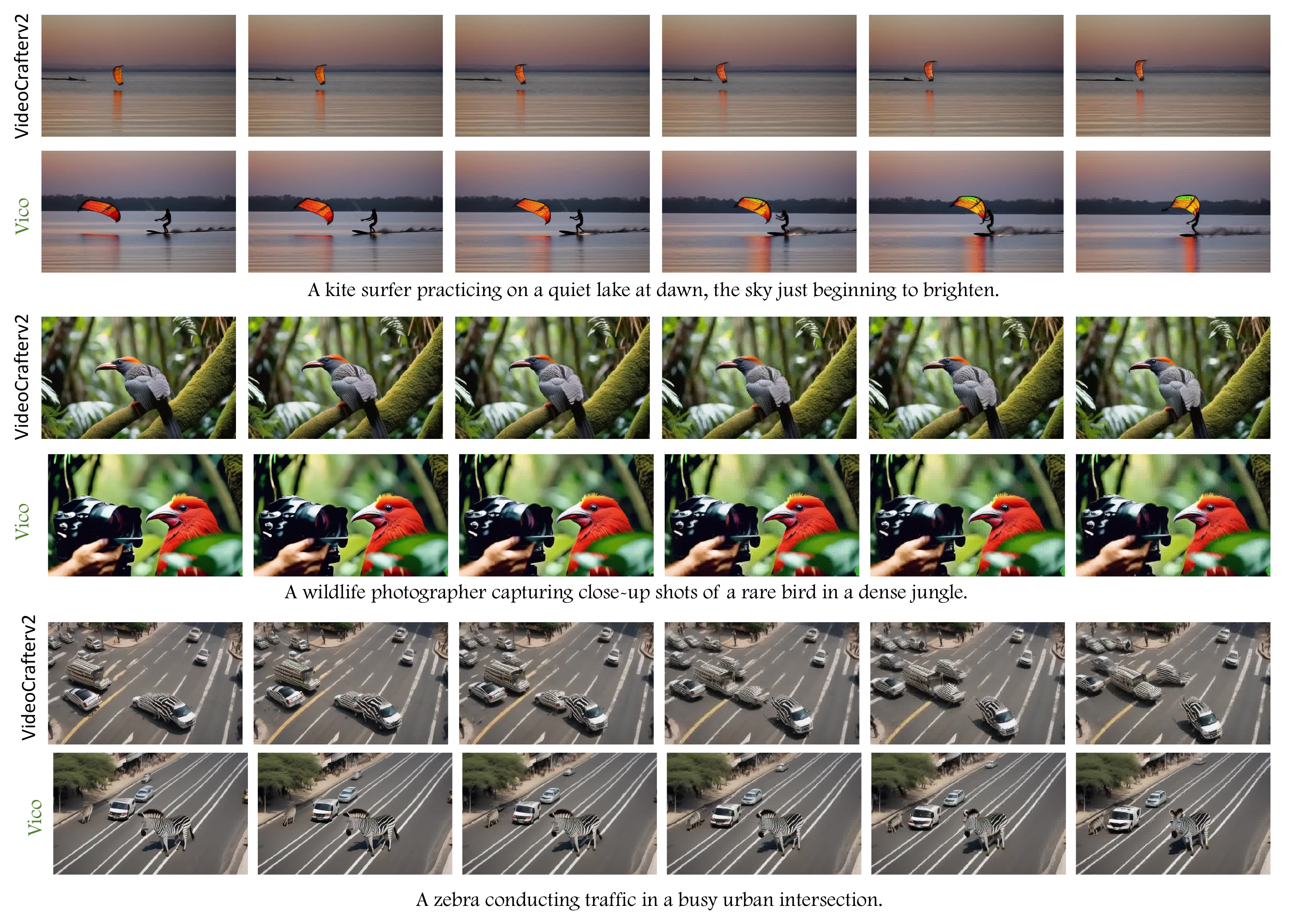}
    \caption{Video visualization for compositional video generation}
    \label{fig:moreresults}
\end{figure}

Here we provide more example for compositional T2V in Figure~\ref{fig:moreresults}
\section{Speed Analysis}
\label{sec:speed}

\noindent\textbf{Attribution Speed.} In this section, we assess the running speed of our ST-flow. To assess its computational efficiency, we compare ST-flow with cross-attention and Attention Rollout~\cite{DBLP:conf/acl/AbnarZ20} computation, by reporting the theoretical complexity and empirical running time. We assume we have  1 cross attention map of $mxn$ and $L$ self-attention map of $n\times n$, and demonstrated the theoretical results. Specifically, we measure the average running time required for each diffusion model inference, focusing solely on the time taken for attribution computation, excluding the overall model inference time. We use the VideoCrafterv2 as the base model.

As detailed in Table~\ref{tab:attriution speed}, the cross-attention computation is fast, as it processes only a single layer. Both Attention Rollout and our approximated ST-Flow involve matrix multiplications and consequently share a similar time complexity. However, our ST-Flow approximation benefits from the relatively faster speed of element-wise min-max operations compared to the floating-point multiplications used in Attention Rollout, leading to slightly quicker execution times.

In contrast, the exact ST-Flow method is much slower. This is because it requires independently estimating the flow for each sink-source pair, a process that takes considerable time.

\begin{table}[H]
    \centering
    \begin{tabular}{l|c|c}
    \toprule
        Method & Complexity & sec/inference  \\
        \midrule
        Cross-Attn. & O(1) & 0.002s\\
        Attention Rollout & $O(Lmn^2)$ & 0.042s\\
        Exact-ST-Flow & $O(L^3mn^4)$ & ~8s\\
        ST-Flow (soft)& $O(Lmn^2)$ & 0.037s\\
        \bottomrule
    \end{tabular}
    \caption{Speed comparison for attribution method.}
    \label{tab:attriution speed}
\end{table}

\noindent\textbf{Diffusion Inference Speed.} Our Vico framework includes a iterative optimization process alongside with the denoising. As expected, it should results in longer inference time. We evaluated this using a 50-step DPM denoising process on the VideoCrafterv2 model, at a resolution of $512\times320$ for 16 frames, on a single A6000 GPU.

The results, shown in Table~\ref{tab:time}, reveal that the baseline VideoCrafterv2 completed in 23 seconds. Adding the Attend\&Excite increased the duration to 48 seconds. In comparison, our Vico framework finished in a comparable time of 50 seconds. Despite its additional complexity, Vico's efficient design keeps the inference time within a reasonable range.

\begin{table}[H]
    \centering
    \begin{tabular}{l|c}
    \toprule
        Method & Time \\
        \midrule
        VideoCrafterv2 & 23s\\
       + Attend\&Excite & 48s \\
        + \texttt{Vico} (soft\&hard)& 45s \\
        \bottomrule
    \end{tabular}
    \caption{Text-to video model inference time comparison.}
    \label{tab:time}
\end{table}

\section{Implementation details of Vico}

\noindent\textbf{ST-Flow Computation.} To compute the ST-Flow, we begin by extracting attention weights from all layers. These weights are averaged across all heads and then upscaled to the image size using bicubic interpolation. Due to the block-wise sparse pattern of the connections, min-max matrix multiplication is applied to the capacity matrix for connected layers. Furthermore, given that cross-attention layers include skip connections from previous layers, we divide the network into multiple groups. Within each group, min-max matrix Multiplication is performed. Finally, we aggregate the scores across all groups to obtain the results. The pseudocode for the min-max multiplication is in~Algorithm~\ref{alg:minmax}.

\begin{algorithm}
\caption{Batched Min-Max Matrix Multiplication}
\label{alg:minmax}
\begin{algorithmic}[1]
\Function{BatchMinMaxMatrixMultiplication}{$A, B$}
    \State \textbf{Input:}
    \State $A$ is a tensor of shape $[B, m, k]$
    \State $B$ is a tensor of shape $[B, k, n]$
    \State \textbf{Output:}
    \State Tensor of shape $[B, m, n]$ containing the maximum values

    \Statex

    \State $A_{\text{expanded}} \gets A.\text{unsqueeze}(2)$ \Comment{Shape becomes $[B, m, 1, k]$}
    \State $B_{\text{expanded}} \gets B.\text{permute}(0, 2, 1).\text{unsqueeze}(1)$ \Comment{Shape becomes $[B, 1, n, k]$}

    \Statex

    \State $min\_vals \gets \text{torch.min}(A_{\text{expanded}}, B_{\text{expanded}})$ \Comment{Shape becomes $[B, m, n, k]$}
    \State $max\_vals \gets \text{torch.max}(min\_vals, \text{dim}=3).\text{values}$ \Comment{Shape becomes $[B, m, n]$}

    \Statex

    \State \Return $max\_vals$
\EndFunction
\end{algorithmic}
\end{algorithm}

\noindent\textbf{Latent Step.} During the first half of the sampling process, we update the latent variables. We establish a loss threshold of 0.2; once this threshold is reached, no further updates are made.

\section{Limitations}
Although Vico effectively allocates attribution across different tokens, it does not explicitly bind attributes to subjects. Moreover, there is a critical balance to maintain between latent updates and semantic coherence. Excessive updating can lead to the generation of nonsensical videos.

\section{Broader Applications}
Technically, the computation of attention flow proposed in our system is versatile and can be efficiently applied to a variety of other applications like erase certain concept in diffusion models. Additionally, the principle of fairly distributing the contribution of different input parts can be extended to other domains, such as language modeling.

\end{document}